\documentclass[letterpaper]{article} 
\usepackage{aaai24}  
\usepackage{times}  
\usepackage{helvet}  
\usepackage{courier}  
\usepackage[hyphens]{url}  
\usepackage{graphicx} 
\usepackage{natbib}  
\usepackage{caption} 
\usepackage{algorithm}

\usepackage{newfloat}
\usepackage{listings}
\DeclareCaptionStyle{ruled}{labelfont=normalfont,labelsep=colon,strut=off} 
\floatstyle{ruled}
\newfloat{listing}{tb}{lst}{}
\floatname{listing}{Listing}
\usepackage{amsmath}
\usepackage{amsfonts}
\usepackage{xfrac}
\usepackage{dsfont}
\usepackage{algorithm}
\usepackage{algpseudocode}
\usepackage{graphicx}
\usepackage{capt-of}
\usepackage{multirow}
\usepackage{subfig}
\usepackage{caption}

\usepackage{amssymb}
\usepackage{mathtools}
\usepackage{amsthm}
\usepackage{todonotes}

\DeclareMathOperator*{\argmax}{argmax}
\DeclareMathOperator*{\argmin}{argmin}

\title{Sample-and-Bound for Non-convex Optimization}

\author{
    Yaoguang Zhai\equalcontrib,
    Zhizhen Qin\equalcontrib,
    Sicun Gao
}

\affiliations{
    University of California, San Diego\\
    \{yazhai, zhizhenqin, sicung\}@ucsd.edu

}

\usepackage{bibentry}

\begin{document}

\maketitle

\begin{abstract}
Standard approaches for global optimization of non-convex functions, such as branch-and-bound, maintain partition trees to systematically prune the domain. The tree size grows exponentially in the number of dimensions. 
We propose new sampling-based methods for non-convex optimization that adapts Monte Carlo Tree Search (MCTS) to improve efficiency. 
Instead of the standard use of visitation count in Upper Confidence Bounds, we utilize numerical overapproximations of the objective as an uncertainty metric, and also take into account of sampled estimates of first-order and second-order information. 
The Monte Carlo tree in our approach avoids the usual fixed combinatorial patterns in growing the tree, and aggressively zooms into the promising regions, while still balancing exploration and exploitation. 
We evaluate the proposed algorithms on high-dimensional non-convex optimization benchmarks against competitive baselines and analyze the effects of the hyper parameters. 

\end{abstract}

\section{Introduction}
Non-convex global optimization problems are pervasive in engineering \cite{mistakidis2013nonconvex, campi2015non}, computer science \cite{liu2014solving, jain2017non}, and  economics \cite{bao2010set}.
The problem is well-known to be NP-hard, and the practical challenge lies in distinguishing the global optimum from exponentially many potential local optima~\cite{jain2017non, yang2019advancing}. 

Existing approaches to non-convex optimization can be largely categorized into sampling-based methods and tree-search methods.
Sampling-based approaches, such as simulated annealing (SA) \cite{simulatedannealing} and cross-entropy (CE) \cite{de2005tutorial},  explore the solution space through random sampling and guide search strategies with the minimal assumptions about the objective function. 
Sampling methods can be designed to asymptotically converge towards the global optimum, but suffer from the curse-of-dimensionality in practice. Tree search and interval-based optimization methods \cite{gurobi_manual, ninin2016global} leverage various branch-and-bound techniques that maintain a partition tree over the domain to systematically prune the space towards global optima. 
Such algorithms typically employ rigorous techniques (e.g., linear relaxation \cite{yanover2006linear} and interval arithmetic \cite{hickey2001interval, araya2016interval}) for bounding the functions and systematically explore the solution space. The size of the search tree can quickly become exponential in the number of dimensions and is the major bottleneck for scaling up. 

We propose an approach that combines the benefits of sampling-based and tree-based approaches as well as interval bounding and local optimization techniques, by taking advantage of the recent progress in Monte Carlo Tree Search (MCTS) methods. 
We assume that the analytic form of the objective function is known over a compact domain, so that we can use interval bounding~\cite{araya2016interval} on the function value and its local first-order and second-order information in different parts of the MCTS design. A key feature of the Monte Carlo trees is that the growth of the tree is driven by samples rather than partitions, and hence the name {\em Sample-and-Bound}. By associating the analytic and estimated properties of adjustable neighborhoods around each sample, we design the MCTS algorithm to best balance exploration and exploitation based on the important numerical properties of the objective function. We evaluate the proposed algorithms against a wide range of existing algorithms and analyze the importance of various hyper parameters.





\section{Related Work}
Some classical approaches to global optimization
explore the search space by sampling without explicitly building models of the objective function. 
Common techniques in this category include stochastic methods such as SA \cite{simulatedannealing} and CE \cite{de2005tutorial}, as well as deterministic schemes like Nelder-Mead (NM) \cite{neldermead}. 
SA \cite{simulatedannealing} uses a probability-driven search process to escape local minima. 
CE \cite{de2005tutorial}, on the other hand, is a technique that iteratively updates the probability distribution on the search space to look for optimal regions.
NM method \cite{neldermead} is a deterministic sampling approach, which maintains a simplex within the search space and updates its vertexes based on evaluations at selected points.

Sampling-based approaches have recently been combined with tree search by building a search tree for the state space and pick only the most promising subspace to sample.
Existing algorithms include Deterministic Optimistic Optimization (DOO) \cite{DOO2011}, Latent Action Monte Carlo Tree Search (LaMCTS) \cite{lamcts2020}, and Monte Carlo Tree Descent (MCTD) \cite{zhaimonte}. 
DOO segments the search domain into sections, each represented by a point; and the new sample is carefully selected by choosing the most suitable section. 
LaMCTS method employs MCTS to manage search space partitioning. 
It learns latent actions to distinguish good and bad regions in the search space, and samples in the good partitions during its tree's expansion. 
MCTD assumes the objective function as black-box, utilizes a combination of sampling based approach and learning based approach for local optimization, and employes MCTS to select the best local optimization processes. 
Although these methods have adeptly laid out a comprehensive framework for navigating the search space, the task of identifying the most promising subspace from the sample data remains a challenge.

Another category of global optimization methods require the access to the formulation and rely on precise anticipation of objective function values within predefined regions. 
They employ the branch-and-bound algorithms in which they constitutes a robust framework that systematically partitions the solution space into more accessible sub-regions referred to as branches. 
The evaluation of each branch is made according to its potential to outperform the current optimal solution based on the bounding of objective function intervals specific to that branch. 
As the algorithm advances, it tactically prunes branches that can be definitively identified as incapable of providing a superior solution.
The typical solvers for this category are BARON \cite{ts:05, baron_manual} and Gurobi \cite{gurobi_manual}.
BARON \cite{ts:05, baron_manual} is explicitly tailored to address non-convex global optimization problems by strategically exploring the solution space. 
Its purpose is to either uncover globally optimal solutions or provide verified lower bounds for the optimal objective value. 
It achieves this through accurate bounding of non-linear terms with several exceptions such as trigonometric functions and min/max functions.
Gurobi \cite{gurobi_manual} is a widely used commercial optimization solver famous for its proficiency in handling quadratic programming problems and various optimization scenarios. 

\section{Preliminary}
\paragraph{Problem Formulation.}
We consider the problem of minimizing an objective function $f(x): \Omega\rightarrow \mathbb{R}$, where the domain $\Omega \subseteq \mathbb{R}^n$ is compact. 
In our approach, we assume that we have access to the analytical form of the function $f(x)$, enabling us to query its first-order derivative vector $G(x) = f'(x)$, the second-order partial derivative Hessian matrix $H(x) = f''(x)$, and evaluate the function value interval $f(B)$ over a specified input box $B \subseteq \Omega $.

\paragraph{Interval Arithmetic.}
Interval computation is a mathematical and computational approach that operates on quantities and variables represented as intervals \cite{alefeld2012introduction}. 
In this context, for a function $f$ defined on an input box domain $B$, the value of $f(B)$ is expressed as an interval $[lb, ub]$, where for every $x$ within $B$, the function value satisfies the inequality $lb \leq f(x) \leq ub$.

\paragraph{Monte Carlo Tree Search.}
MCTS effectively balances exploration and exploitation based on the theory of multi-armed bandits. 
The MCTS framework consists of four main steps: Selection, Expansion, Simulation, and Backpropagation.
During the Selection step, the search tree is traversed from the root node to a leaf node. 
The Upper Confidence Bound for Trees (UCT) value, defined as Eq.~\ref{eq:uct_classical}, is used to select the best child of a parent node:

\begin{equation}
\label{eq:uct_classical}
\nu(n_i) = \frac{R_i}{N_i} + C \cdot \sqrt{\frac{2 \cdot \ln(N_p)}{N_i}}
\end{equation}
Here, $R_i$ represents the rewards obtained on child node $n_i$, $N_i$ is the number of visits to $n_i$, $N_p$ is the number of visits to $n_i$'s parent node $n_p$, and $C$ is a constant that balances exploration and exploitation. From the root node, the algorithm recursively select the child node with the highest UCT value, until a leaf node is reached.
During the Expansion step, a new child node is added to the selected leaf node. 
In the Simulation step, a random simulation is performed from the newly added child node until a terminal node is reached, and the simulation reward is estimated.
Finally, in the Backpropagation step, the simulation reward is propagated backward from the expanded node to the root node, whose statistics are updated accordingly.

\section{Monte Carlo Tree Search with Interval Bounds and Regional Estimation}

\paragraph{Overview.}

\begin{figure}
    \centering
    \includegraphics[width=.45\textwidth]{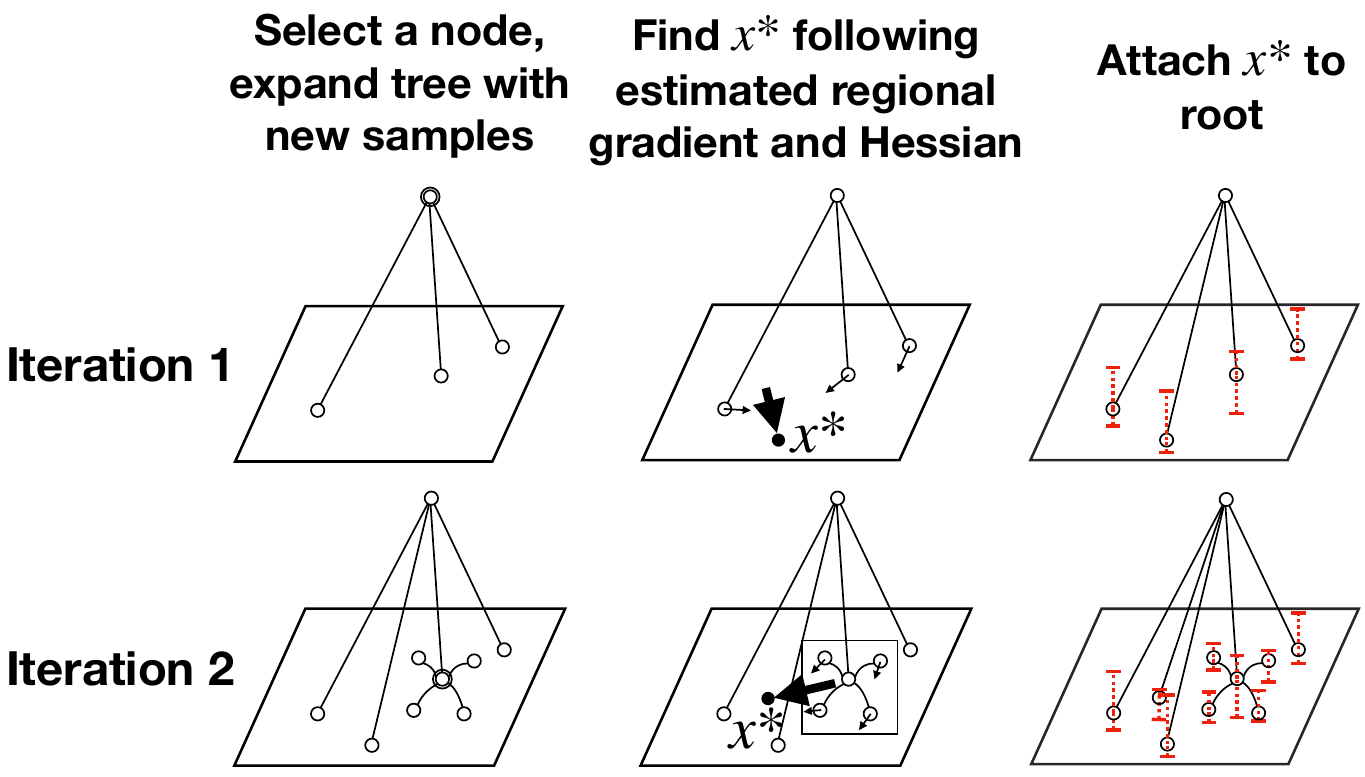}
    \caption{Steps in each iteration of the MCIR algorithm}
    \label{fig:tree_flow}
\end{figure}

The pseudocode of MCIR is provided in Alg.~\ref{alg:MCIR}, and a high-level visualization is depicted in Fig.~\ref{fig:tree_flow}.
Our MCIR algorithm employs a search tree structure constructed based on collected samples of the objective function and follows a systematic searching approach in each iteration. 
Each node in the tree contains a batch of samples encompassed within a box domain associated with that node.
In every iteration, we select a leaf node $n_p$ using a modified UCT formula with function evaluation on the box, and we expand the selected node by adding new child nodes $n_{ci}$ generated from random sampling (Fig.~\ref{fig:tree_flow} (a)).
We also identify another new child node $n^*$ by leveraging the regional estimation based on gradient and Hessian within the neighborhood of the selected node $n_p$ (Fig.~\ref{fig:tree_flow} (b)). 
The node $n^*$ represents a node superior to the selected $n_p$, and we attach it with the root node.
This attachment allows us to prioritize the search on this node, thereby accelerating the search process (Fig.~\ref{fig:tree_flow} (c)).
For each newly created node we run local optimization with limited steps to enhance the quality of the best-found sample on it. 

Note that despite the special design of different parts of MCTS for the optimization context, the proposed algorithm ensures non-zero probability of sampling any neighborhood with positive measure in the input space. 
Consequently, MCIR is complete in the sense that it will eventually find an $\varepsilon$-neighborhood around the optimal value of any continuous function with arbitrarily small positive $\varepsilon$. 

\begin{algorithm*}[t]
  \caption{Monte Carlo Tree Search with Interval Bounds and Regional Estimation (MCIR)}
  \label{alg:MCIR}
\begin{minipage}[t]{0.53\textwidth}
\begin{algorithmic}[1]
\Function{MCIR}{objective: $f$, domain: $\Omega$}
\State $n_{0} \gets None $  \Comment{root without sample}
\State $B_{0} \gets \Omega$, $lb_{0} \gets f(\Omega), V_{0} \gets V(B_{0})$ 
\For{step = $1, ..., t$}
\State $n_p$ $\gets$ Select($n_0$)
\State EXPAND($n_p$)
\State LEARN($n_p$)
\State BACKUP($n_p$)
\EndFor 
\State \textbf{return} $ y^*_{0}$ 
\EndFunction
\State

\Function{Select}{node: $n$}
\label{alg:Select}
\State $n_p \gets n$
\While{$n_p$ has children}
\For{$n_{ci} \in n_p$.children}
  \State $u(n_{ci}) \gets u(y^*_{ci}, lb_{ci}, V_{ci})$ by Eq.~\ref{eq:uct}
\EndFor

\State $j$ $\gets$ $ \argmax_{i} {u(n_{bi})} $
\State $n_{p} \gets n_{cj}$
\EndWhile
\State \textbf{return} $n_p$
\EndFunction

\State

\Function{Backup}{node: $n_p$}
\label{alg:BackPropagate}
\If{$n_p$ has children}
\State $\{ n_{ci}\} \gets n_p$.children
\State $y^*_p = \min_{i} y^*_{ci}$
\State $j = \argmin_{i} lb_{ci}$
\State $lb_{p} = lb_{cj}$
\State $V_{p} = V_{cj}$
\EndIf
\State BACKUP($n_p$.parent)
\EndFunction

\end{algorithmic}
\end{minipage}
\begin{minipage}[t]{0.46\textwidth}
\begin{algorithmic}[1]

\Function{Expand}{node: $n_p$}
\label{alg:Expand}
\State $B = B_p $
\While{$ B \neq \emptyset $}
\State $x_{ci} \gets x\in B$
\State $x_{ci} \gets$ LocalOpt $(x_{ci}, B_{ci})$
\State $n_{ci} \gets x_{ci}\in B$
\State $B_{ci} \gets b \in B, x_{ci} \in b$
\State $lb_{ci} \gets f(B_{ci})$
\State $V_{ci} \gets V(B_{ci})$
\State $B \gets B \backslash B_{ci}$ 
\State $n_{p}$.children.append($n_{ci}$)
\EndWhile
\EndFunction

\State 
\Function{LEARN}{node: $n_p$}
\label{alg:Learn}
\State $n_{ci} \gets n_p$.children
\State $\overline{H} \gets \overline{Hessian(x_{ci})} , i=1, ...$
\State $j$ $\gets$ $ \argmin_{i} {(y_{bi})} $

\State $\{ x'\} \gets x$, for $|x - x_{cj} | < \delta$

\State $\overline{G} = \overline{grad(x')}$

\For{$ d = 1, ..., dims $}
\If {$\overline{H}_{dd} > 0$}
\State $x^*_{d} = x_{cj, d} - \overline{G_{d}} / \overline{H_{dd}} $
\Else
\State $x^*_{d} = x_{cj, d} - \overline{G_{d}} $
\EndIf
\EndFor
\State $n^* \gets x^*$
\State $B^* \gets B_{cj}$, centered at $x^*$
\State $lb^* \gets f(B^*), V^* \gets V(B^*)$
\State $x^* \gets$ LocalOpt $(x^*, B^*)$
\State $n_0$.children.append($n^*$)
\EndFunction
\end{algorithmic}
\end{minipage}
\end{algorithm*}

\paragraph{Sub-domain Marking.}

Samples are the primary information in each node of the search tree that our algorithm build. Around each sample, we mark up the subdomain around it that is considered at the node.  
The subdomain, typically a hyperbox, will be the focus of local search and optimization at the node, for determining the value of a node. We use the notation $B_{i}$ to denote the box subdomain associated with the node $n_i$.
In the first iteration, the root node $n_{root}$ encompasses the entire search space, $B_{root} = \Omega$, with the function's lower bound on $B_{root}$ denoted as $lb_{root} = lb(f(B_{root}))$, and its box volume $V_{root}$ represented in logarithmic scale.
For subsequent iterations, box $B_{i}$ is assigned to a node $n_i$, while $lb_{i}$ and $V_{i}$ will be updated according to formulas to be described below.
To compute the lower bound of the objective function within a specified input box domain efficiently using global interval bounding~\cite{ninin2016global}.

\paragraph{Path Selection.}

The key to our design is a modified UCT formula that considers both exploration and exploitation. The pseudocode of this procedure can be found in the \textbf{SELECT} function in Alg.~\ref{alg:MCIR}.
For each child node $n_{ci}$ with $i = 1, ...$, and its parent node $n_{p}$, the UCT value $u(n_{ci})$ is determined by the following equation:
\begin{equation}
\label{eq:uct}
u(n_{ci}) = -y^*_{ci} 
       - C_{lb} \cdot lb_{ci} 
       - C_{v} \cdot V_{ci}
       + C_{x}  \cdot \sqrt{\frac{\log{N_{p}}}{N_{ci}}}
\end{equation}
In this formula, $C_{lb}$, $C_{v}$ and $C_{x}$ are weights for the function's lower bound, the volume of the box, and visitation-based exploration, respectively.
The variables $N_{p}$ and $N_{ci}$ denote the number of visits to the parent node $n_{p}$ and the child node $n_{ci}$.
$y^*_{ci}$ indicates the current best function value discovered on the node $n_{ci}$, and $lb_{ci}$ corresponds to the lower bound of the function's interval value on the node $n_{ci}$. 
The term $V_{ci}$ is the volume (in logarithmic scale) of the box where the lower bound is identified. 
It is worth noting that after the creation of new child nodes, the function lower bound $lb_{p}$ and the box volume $V_{p}$ on the parent node $n_p$ can be updated, as detailed in the subsequent section.

This formulation takes into account the following factors to balance exploration and exploitation:
(1) the best function value observed within the box domain, 
(2) the lower bound of the function value within the domain from interval computation, which reflects the potential best function value upon further exploitation, 
(3) the volume of the box where the lower bound is determined, related to the reliability of the function lower bound prediction, 
and (4) the frequency of node visitation. 
While we considered other ingredients - such as upper function value bound, or values from leveraging the function's analytical form - to put into the formula, the design in Eq.~\ref{eq:uct} turns out to be the most effective.

Utilizing Eq.~\ref{eq:uct}, our algorithm tends to redirect its attention to probe alternative sub-domains when a local optimum is identified.
When a box is tight enough, the variance of the objective function in the box is low, so the identified local optimum within the box $lb_{ci}$ is relatively accurate.

If this $lb_{ci}$ is close to the minimum of all other $lb_{ci'}$, it indicates a near-optimal solution has been identified. 
Conversely, if an $lb_{ci'}$ exists that is substantially lower than the current $lb_{ci}$, the search scheme leans towards selecting the node with the lower $lb_{ci'}$ value in the subsequent iteration due to the path selection criterion Eq.~\ref{eq:uct}.
In summary, Eq.~\ref{eq:uct} within our algorithm helps strike a balance between exploiting the neighbor region around the current best-found point and exploring other domains that might contain lower function values.

\paragraph{Tree Expansion.}
\label{chap:tree_exp}
In our algorithm, we utilize two steps to expand the tree effectively. 
The first step involves sampling within the box of the parent node and generating new child nodes based on these chosen samples. 
The second step emphasizes learning a high-quality sample point by leveraging both global Hessian and local gradients.

After selecting the leaf node, we proceed to exploit the function space by sampling and creating a cluster of child nodes within the corresponding box (\textbf{EXPAND} in Alg.~\ref{alg:MCIR}). 
To ensure comprehensive coverage, we divide the box $B_{p}$ into distinct subsets $B_{ci}$ for each child node $n_{ci}$, satisfying  $\cup \{ B_{ci} \} = B_{p} $ and $B_{ci} \cap B_{cj} = \emptyset, i \neq j$. 
Additionally, local optimization may be applied to each individual child node $n_{ci}$ to improve sample quality. 
When a child node $n_{ci}$ is created, we ascertain its function lower bound $lb_{ci}$ through interval propagation of the corresponding box $B_{ci}$: $lb_{ci} = lb(f(B_{ci}))$. 
Once the cluster of child nodes is created and their boxes fully cover the parent box $B_{p}$, we update the lower bound on the parent node $lb_{p} = \min (lb_{ci}) = lb_{cj}$ and the volume of the associated box $V_{p} = V_{cj} $, where $i = 1, .., j, ...$. 
This update will be propagated to the root node.

Next, we learn a representative node $n^*$ using the current set of samples $n_{ci}$ from the selected node $n_p$, as outlined in Alg.~\ref{alg:MCIR} \textbf{LEARN}. 
This step is performed by computing the diagonal of the Hessian matrix, $diag(H)$, for each child node $n_{ci}$, and estimating the expected value. 
Furthermore, we collect the gradient information $G$ around the best sample of $n_{ci}, i=1,...$ and perform a step of Newton's method (or gradient decent when Newton's method is not conductive to minimization), starting from the best sample. 
The average Hessian, derived from the broad region within the box, represents the overall curvature characteristics of the box. 
By integrating locally-averaged gradient information, we can identify a sub-region within the box that is more likely to encompass a minimum.
The learned representative node, denoted as $n^*$, is attached to the root node.
Note that this attachment means the root node can have children nodes $n_{ci}$ and $n_{cj}$ where $B_{ci} \cap B_{cj} \neq \emptyset$.
This step grants $n^*$ higher priority in subsequent iterations. 
Such prioritization promises to guide the search toward a favorable region, thereby reducing unnecessary tree expansion and preserving tree manageability. Considering that this step may expand the tree's first level of children in every iteration, an extra step may be taken to evaluate the quality of the newly learned node and prune unnecessary ones. 

\paragraph{Local Optimization.}
Upon creating a child node, we have the option to conduct local optimization steps to improve the quality of the samples on the node.
While this step is not obligatory, it offers a beneficial opportunity to refine the samples on each node.
To ensure efficient execution, the number of optimization steps is typically kept at a low value, preventing over-exploitation of the immediate local neighborhood.
Local optimization can utilize a variety of numerical optimization algorithms. 
Since the representative node has already been learned using second-order information, we make quasi-Newton methods such as L-BFGS-B~\cite{byrd1995limited, zhu1997algorithm} the default choice for local optimization. 
To ensure computational efficiency, we limit the number of function evaluations during the local optimization. 
In most cases we cap the number of iterations at fewer than $50$, as we do not want to overemphasize the choice of the local optimizer.
It is worth mentioning that alternative local optimization algorithms can be employed based on specific requirements and preferences.

\paragraph{Backward Propagation.} After creating and locally optimizing children nodes $n_{ci}$ and $n^*$, we back propagate three important values upwards as in Alg.~\ref{alg:MCIR} \textbf{BACKUP}, to enhance efficient exploration and decision-making in the subsequent steps.

Firstly, we update the best function value $y^*_{ci}$ found on the child node $n_{ci}$, to the parent node $n_{p}$ with $y^*_{p}$. 
This ensures that the parent node retains the most optimal function value discovered within its subtree.
Secondly, we update the lower bound of the function interval value $lb_{p}$ on the parent node $n_{p}$ with $lb_{p} = \min (lb_{ci})$. 
Given that the newly created child nodes comprehensively cover the box of the parent node, this update provides more precise information guiding the search towards the global minimum.
Lastly, we propagate the size of the box $V_{ci}$ from which the lower bound of the function value originates: $V_{p} = V_{cj}$, where $j = \argmin_{i} lb_{ci}$. 
This box size represents the uncertainty in the input search space concerning the approximated function interval value. 
The same propagation is applied to the node $n^*$, even though its parent is the root node.



\section{Experiments}

\paragraph{Benchmarks.}
To evaluate the performance of our algorithms, our benchmark sets include three distinct categories: synthetic functions designed for nonlinear optimization, bound-constrained non-convex global optimization problems derived from real-world scenarios, and neural networks fitted for single valued functions. 
It is important to note that our approach relies on having access to the symbolic expression of the objective function and do not consider other relational constraints to the variables (e.g., "$<=$").
As a result, benchmark sets that are commonly used for black-box optimization problems and constraint optimization problems are not applicable in our case. 

Synthetic functions are widely-used in nonlinear optimization benchmarks \cite{nonlinearbenchmark}.
These functions usually have numerous local minima, valleys, and ridges in their landscapes which is hard for normal optimization algorithms. 
In our tests, we choose three functions: Levy, Ackley, and Michalewicz, and examine our algorithm's performances on the functions in 50d, 100d, and 200d. 
For our evaluation of non-convex global optimization problems in various fields, we select bound-constrained problems from the collection presented in \cite{baron_dev, puranik2017bounds} that do not involve any additional inequality or equality constraints.
To strike a balance between computational resources and the complexity of the function landscapes, we specifically select functions with input dimensions between $30$ and $1000$, and ensure that the functions could be evaluated within a reasonable time, considering the computational cost of computing the gradient and Hessian. 
The chosen functions for our evaluation include Biggsbi1 (1000d), Harkerp (100d), and Watson (31d). 
It is worth noting that this set of test functions is also utilized in the development of BARON \cite{ts:05} and continues to be used in the latest version \cite{baron_manual}.
In addition to the aforementioned problems, we also explore the application of one-layer neural networks with ReLU activation functions fitted for specific objective functions. 
The nonlinearity introduced by activation functions and the partitioning of the input space pose challenges in finding the global minimum of neural networks.
To assess the performance of our algorithm, we train neural networks with varying numbers of input dimensions and one layer of 16 hidden unite.
We translate the entire network into an analytic expression form, enabling us to evaluate the algorithm's effectiveness in optimizing neural network models.
We conduct our experiments on a local machine with Intel(R) Core(TM) i7-8700 CPU @ 3.20GHz, 16G RAM, and NVIDIA GeForce GTX 1080 graphic card.

\paragraph{Baselines.}
We select various sampling-based global optimization algorithms as baselines for our experiments, including: basinhopping \cite{olson2012basin}, differential evolution \cite{storn1997differential,pant2020differential}, dual annealing \cite{xiang2013generalized}, direct \cite{gablonsky2000locally}, CMA \cite{nikolaus_hansen_2023_7573532}, TuRBO \cite{eriksson2019scalable}, LaMCTS \cite{wang2020learning}, and Gurobi \cite{gurobi_manual}.
It should be pointed out that some algorithms, including TuRBO and LaMCTS, are GPU-ready. 
However, due to the limitations of our computational resources, we refrain from using GPUs for the optimization process, except for tasks related to training and evaluating the neural network model.
To compensate for the reduced performance from utilizing the CPU, we extend the timeout for TuRBO and LaMCTS to be five times of other baselines.

It is important to mention that we do not incorporate BARON \cite{baron_manual} as one of our baseline methods, despite its renowned ability to efficiently bound boxes. 
The reason behind this decision lies in the fact that BARON can manage the functions present in their test sets during pre-processing, entirely eliminating the need to execute the optimization algorithm. 
For instances like Biggsbi1, Harkerp, and Watson, BARON can solve them instantly, requiring zero seconds and iterations. 
Moreover, BARON encounters challenges with certain function types, including but not limited to trigonometric functions and min/max functions \cite{baron_manual}. 
These types of functions are prevalent in synthetic test function sets as well as function sets based on neural networks.

Another consideration is that Gurobi requires expertise and extra effort to achieve peak performance. 
While Gurobi stands out as an exceptionally efficient and versatile optimization solver, especially in the context of non-convex optimization problems, it comes with certain prerequisites. 
Its handling of non-linear terms, for instance, treats them as General Constraints, which necessitates extra manual modification to the objective function expression, as outlined in \cite{gurobi_manual}. 
This specific trait might limit our ability to deploy it on entire test sets.


\paragraph{Metrics.}
For each benchmark function, we conduct experiments using baseline algorithms and our proposed algorithm with five different random seeds.
The time limits for the baselines are set to 2 hour. 
Due to CPU utilization, the limits for TuRBO and LaMCTS are extended to 10 hours.
Throughout the experiments, we track the best-found function value until each step and compute the mean and standard deviation across all runs.
This allows us to compare the final best-found values as well as the speed at which each algorithm converges to the optimal result.


\paragraph{Overall Performance.}
\begin{figure*}[t]
\centering    
\includegraphics[width=.92\textwidth]{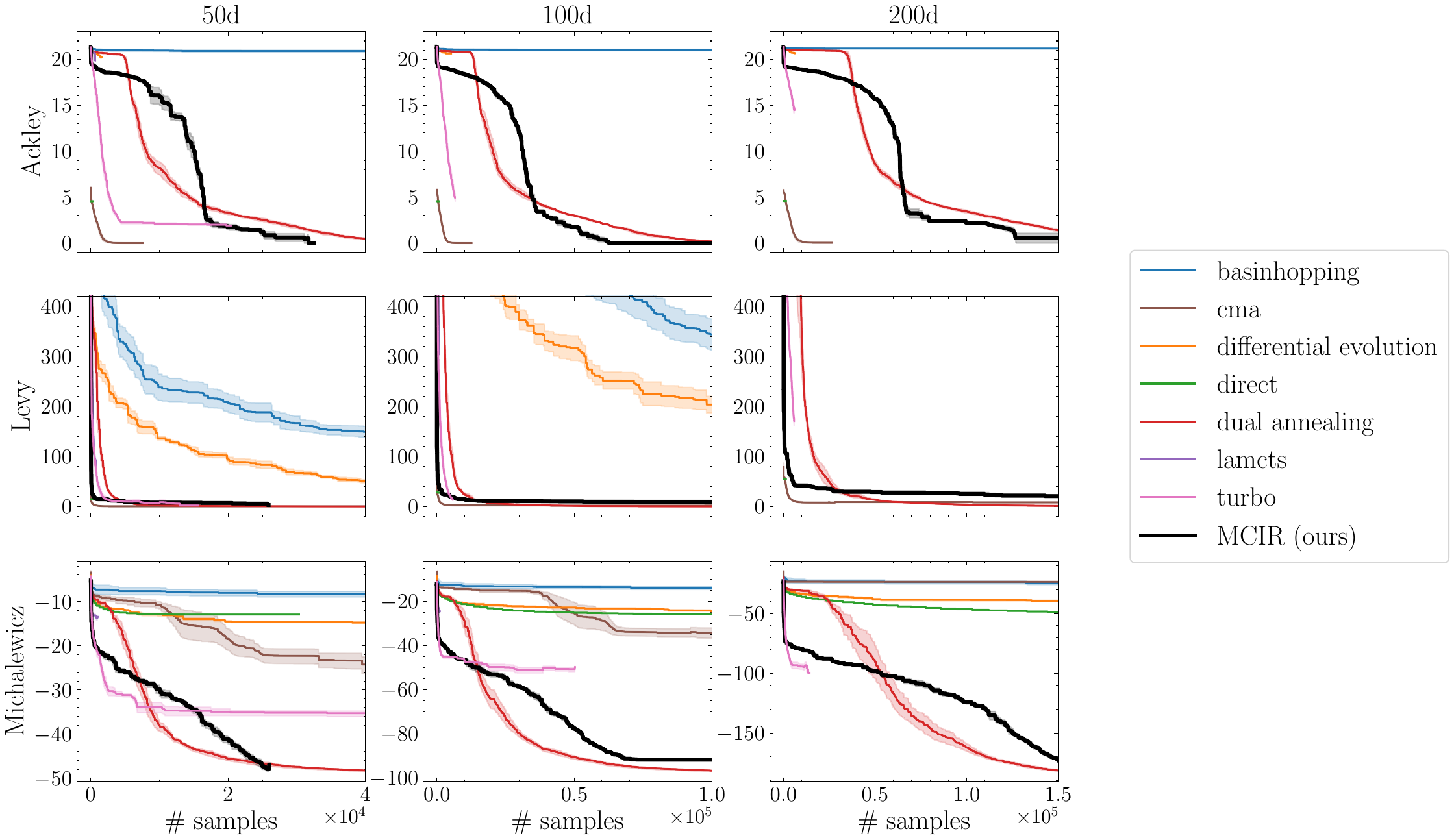}
\caption{Overall performance of the baselines and MCIR on tested synthetic functions.}
\label{fig:overallperform_syn}
\end{figure*}

\begin{figure*}[t]
\centering    
\includegraphics[width=0.92\textwidth]{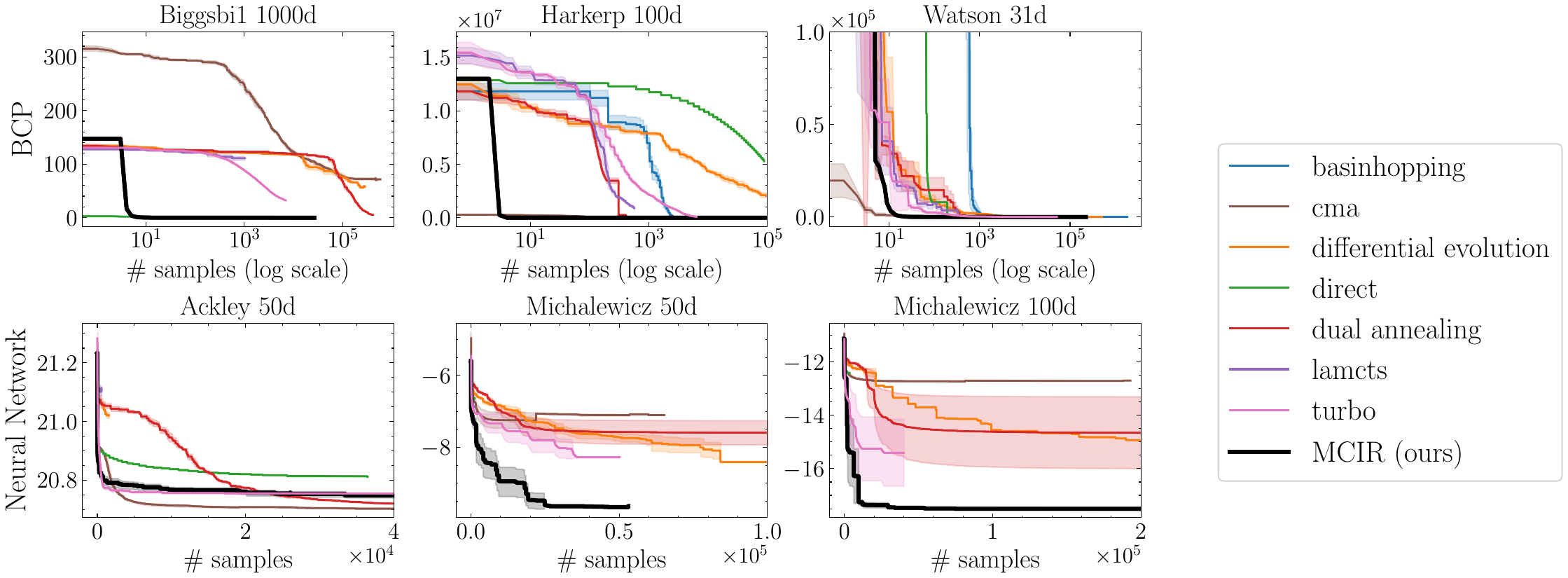}
\caption{Overall performance of the baselines and MCIR on Biggsbi1, Harkerp, Watson, and Neural Networks.}
\label{fig:overallperform_nn}
\end{figure*}


Fig.~\ref{fig:overallperform_syn} presents performance comparisons between the MCIR algorithm and baseline algorithms on the synthetic function benchmark set. 
For the Ackley function (Fig.~\ref{fig:overallperform_syn} first row) and Levy function (Fig.~\ref{fig:overallperform_syn} second row), CMA emerges as the top-performing algorithm, followed by MCIR and dual annealing. 
For the Michalewicz function (Fig.~\ref{fig:overallperform_syn} third row), dual annealing and MCIR delivers similar best performance in terms of final optimization result, while CMA fails to optimize efficiently.
Notably, Gurobi consistently completes its optimization with just one function evaluation call. 
It attains a best function value of $2.02$ on Ackley (for three dimensions), $0.0$ for Levy in both 50d and 100d, and $-0.00016$ for Levy-200d (an anomalous value, because Levy function is greater or equal to $0.0$). 
It also attains the lowest value amongst all algorithms on the Michalewicz function. 

Fig.~\ref{fig:overallperform_nn} offers an in-depth comparison between the MCIR algorithm and the baseline algorithms across benchmarks such as Biggsbi1, Harkerp, Watson, and Neural Networks. 
In the bound-constrained optimization problems (BCP) of Biggsbi1, Harkerp, and Watson (Fig.~\ref{fig:overallperform_nn} first row), MCIR exhibits exemplary performance. 
It adeptly strikes a balance between exploring the search space and executing local optimization, culminating in the precise pinpointing of the global minimum from many suboptimal local minima derived from local optimization. 
CMA shows a performance comparable to MCIR on Harkerp and Watson, and direct algorithm mirrors MCIR's efficacy on Biggsbi1. 
Turning our attention to trained neural networks (Fig.~\ref{fig:overallperform_nn} second row), CMA shines on Ackley-50d, yet MCIR continues to deliver impressive results. 
Notably, for Michalewicz-50d and Michalewicz-100d, MCIR outperforms all other baseline algorithms.

Upon a closer examination of the result curves, it becomes evident that the MCIR algorithm's optimization performance is both commendable and in line with our initial expectations.

\paragraph{Ablation Study.}

\begin{figure*}[t!]
    \centering   
    \subfloat[Michalewicz-50d \\ number of nodes at \\ expansion]{
        \label{fig:nodeexp}
        \includegraphics[width=0.275\textwidth]{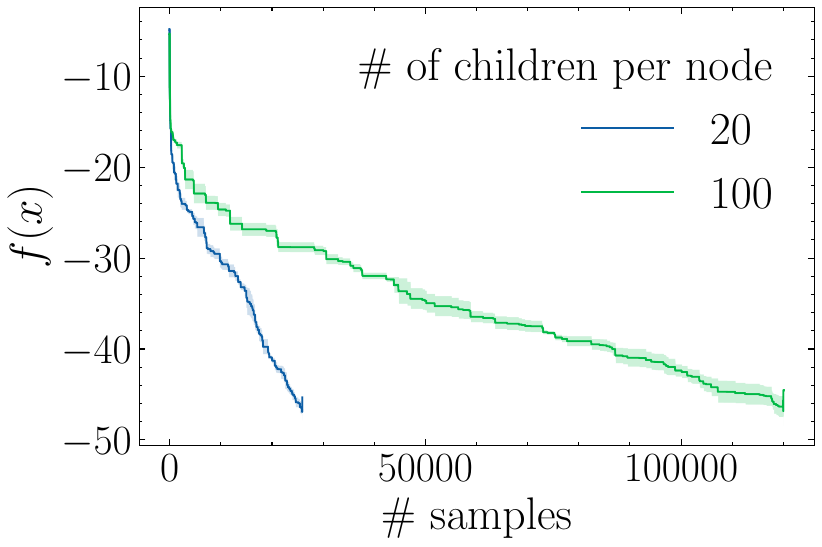}
    }
    \subfloat[Michalewicz-50d \\ number of local  \\ optimization steps]{
        \label{fig:local_micha}\includegraphics[width=0.275\textwidth]{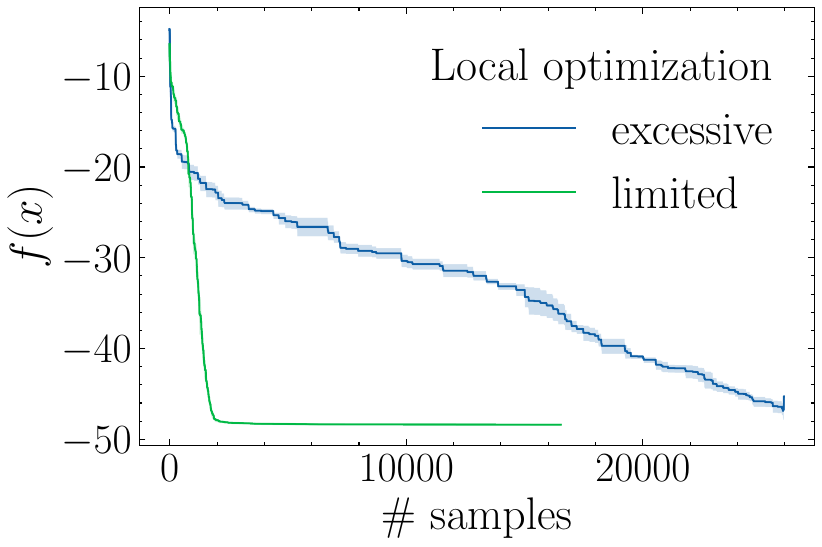}
    }
    \subfloat[Watson-31d \\ number of local \\ optimization steps]{
        \label{fig:local_watson}\includegraphics[width=0.275\textwidth]{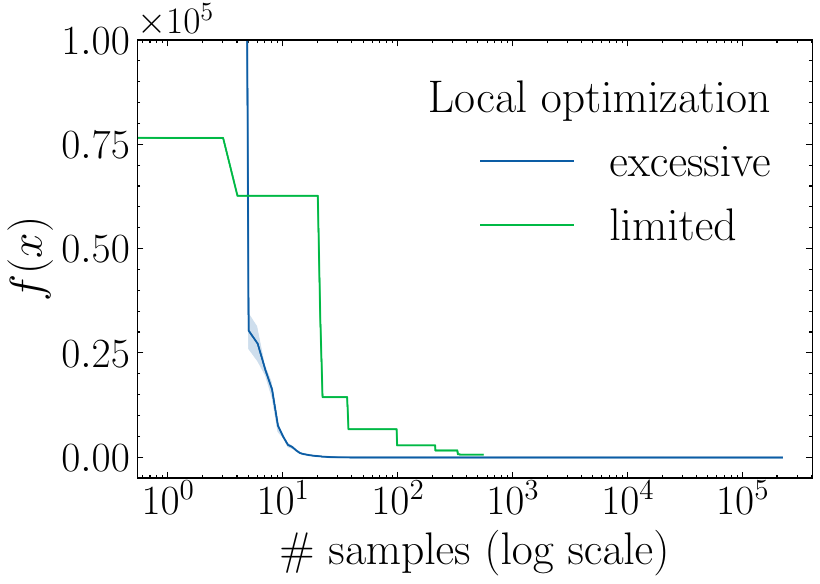}
    }\\
    \subfloat[Ackley-50d, $C_{lb}$]{
        \label{fig:lb}\includegraphics[width=0.283\textwidth]{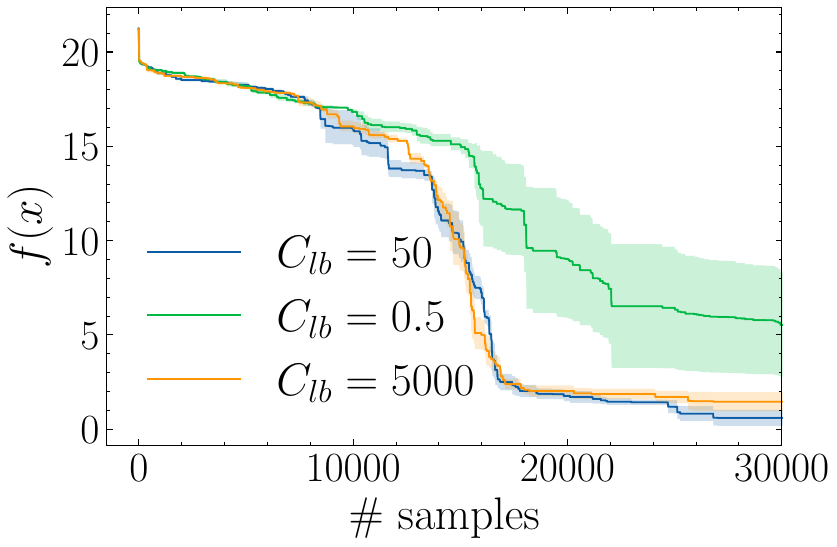}
    }
    \subfloat[Ackley-50d, $C_{v}$]{
        \label{fig:v}\includegraphics[width=0.283\textwidth]{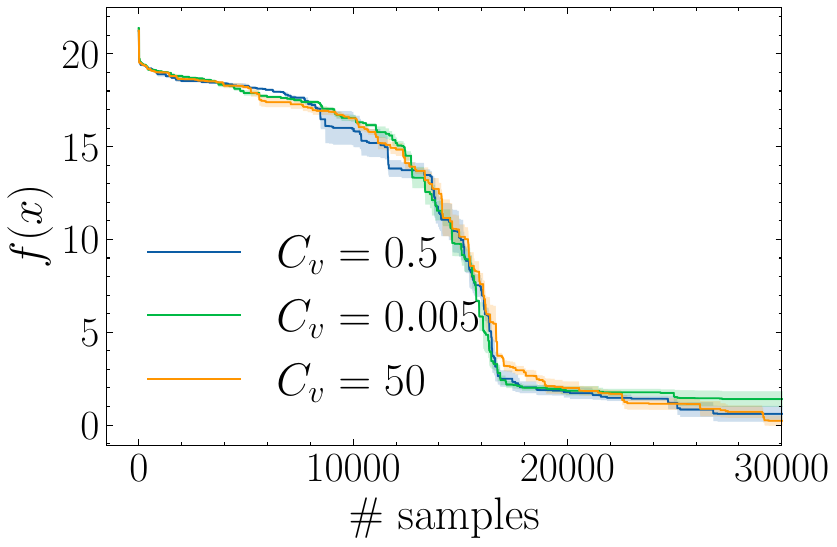}
    }
    \subfloat[Ackley-50d, $C_{x}$]{
        \label{fig:explore}\includegraphics[width=0.283\textwidth]{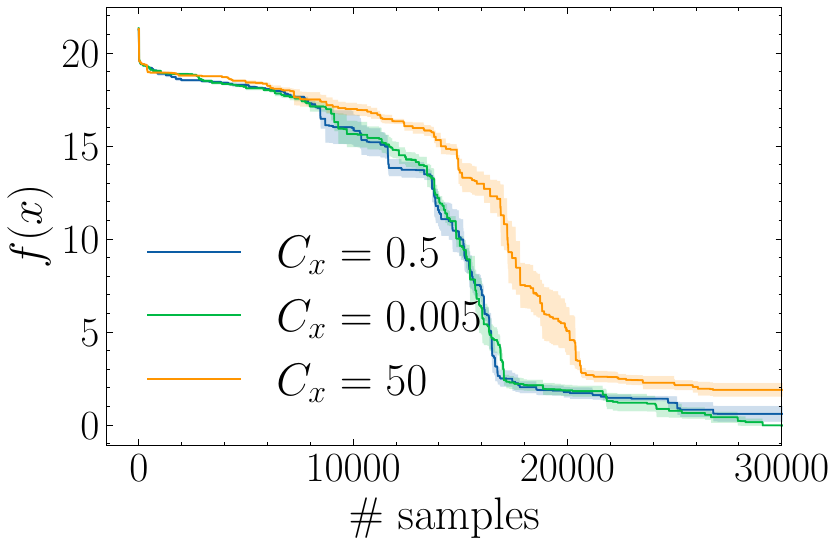}
    }
\caption{Ablation studies on function Michalewicz-50d for the number of children at expansion (a), the effectiveness of local optimization (b), on function Watson-31d for the effectiveness of local optimizaiton (c), and on function Ackley-50d for $C_{lb}$ (d), $C_{v}$ (e), and $C_{x}$ (f).}
\label{fig:ablationstudy}
\end{figure*}

We conducted ablation studies to analyze the individual contributions of different components in our algorithm. 
Specifically, we investigated the influence of the number of random samples placed under each selected node on the Michalewicz-50d function (Fig.~\ref{fig:nodeexp}), assessed the effectiveness of local optimization on Michalewicz-50d (Fig.~\ref{fig:local_micha}), and examined the effectiveness of local optimization on the Watson-31d (Fig.~\ref{fig:local_watson}). 
Furthermore, we tested the hyper parameters used in the UCT formula (Eq.~\ref{eq:uct}) using the Ackley-50d function as depicted in Fig.~\ref{fig:lb}, \ref{fig:v}, and \ref{fig:explore}.
From Fig.~\ref{fig:nodeexp}, we observed that the number of new nodes placed after selecting a leaf node should be kept at a moderate level. 
Overpopulating the same local region with new nodes does not significantly enhance performance due to the closeness of local optima. 
The significance of local optimization can be found in Fig.~\ref{fig:local_micha} and Fig.~\ref{fig:local_watson}. 
It can be concluded that local optimization can both improve and hinder the performance of the algorithm, as shown Fig.~\ref{fig:local_watson} and Fig.~\ref{fig:local_micha}, respectively.
Therefore, setting a reasonable budget for the local optimizer is important. 
A local optimizer with stopping criterion such as improvement threshold could be more preferable over one with fixed number of iterations.
Fig.~\ref{fig:lb}, Fig.~\ref{fig:v}, and Fig.~\ref{fig:explore} demonstrate the importance of the lower bound of the function value in determining the best node for searching the global minimum. 
Additionally, the balance between exploiting nodes excessively and leaving nodes unexplored becomes evident. 
The size of the box where the lower bound originates is the least sensitive hyper parameter, as a smaller box increases the certainty of the function's lower bound but has less impact compared to the value of the function's lower bound itself. 

We have noticed that sometimes the effectiveness of $C_v$ is limited, but as $C_v$ represents the confidence of the predicted objective function interval, the choice of $C_v$ highly depends on the landscape of the objective function. 
For functions where the bounds are evaluated through rough approximation, $C_v$ becomes important.  
In our benchmark tests, we select our hyper parameters $C_{lb}$, $C_v$, and $C_x$ that perform consistently well over multiple tests without requiring too much fine-tuning. 
Most of the parameters are shared across all problems based on the dimension and the type of problems.
In practice, one can start with the setup that provides the best performance in this paper, and fine tune to specific tasks based on observed function complexity and landscape.

\section{Conclusion}
We introduced a new approach to non-convex optimizations problems by leveraging analytic and sampling-based information in an MCTS framework, enabling efficient exploration and exploitation of the state space. 
Experiments results on standard benchmark problem sets demonstrated clear benefits of the proposed approach. 
Future work can focus on reducing the overhead of various numerical computation involved in the proposed algorithm and further optimizing the search tree.

\clearpage

\begin{thebibliography}{33}
\providecommand{\natexlab}[1]{#1}

\bibitem[{Alefeld and Herzberger(2012)}]{alefeld2012introduction}
Alefeld, G.; and Herzberger, J. 2012.
\newblock \emph{Introduction to interval computation}.
\newblock Academic press.

\bibitem[{Araya and Reyes(2016)}]{araya2016interval}
Araya, I.; and Reyes, V. 2016.
\newblock Interval branch-and-bound algorithms for optimization and constraint satisfaction: a survey and prospects.
\newblock \emph{Journal of Global Optimization}, 65: 837--866.

\bibitem[{Bao and Mordukhovich(2010)}]{bao2010set}
Bao, T.~Q.; and Mordukhovich, B.~S. 2010.
\newblock Set-valued optimization in welfare economics.
\newblock \emph{Advances in mathematical economics}, 113--153.

\bibitem[{Byrd et~al.(1995)Byrd, Lu, Nocedal, and Zhu}]{byrd1995limited}
Byrd, R.~H.; Lu, P.; Nocedal, J.; and Zhu, C. 1995.
\newblock A limited memory algorithm for bound constrained optimization.
\newblock \emph{SIAM Journal on scientific computing}, 16(5): 1190--1208.

\bibitem[{Campi, Garatti, and Ramponi(2015)}]{campi2015non}
Campi, M.~C.; Garatti, S.; and Ramponi, F.~A. 2015.
\newblock Non-convex scenario optimization with application to system identification.
\newblock In \emph{2015 54th IEEE Conference on Decision and Control (CDC)}, 4023--4028. IEEE.

\bibitem[{De~Boer et~al.(2005)De~Boer, Kroese, Mannor, and Rubinstein}]{de2005tutorial}
De~Boer, P.-T.; Kroese, D.~P.; Mannor, S.; and Rubinstein, R.~Y. 2005.
\newblock A Tutorial on the Cross-Entropy Method.
\newblock \emph{Annals of operations research}, 134(1): 19--67.

\bibitem[{Eriksson et~al.(2019)Eriksson, Pearce, Gardner, Turner, and Poloczek}]{eriksson2019scalable}
Eriksson, D.; Pearce, M.; Gardner, J.; Turner, R.~D.; and Poloczek, M. 2019.
\newblock Scalable Global Optimization via Local {Bayesian} Optimization.
\newblock In \emph{Advances in Neural Information Processing Systems}, 5496--5507.

\bibitem[{Gablonsky and Kelley(2000)}]{gablonsky2000locally}
Gablonsky, J.~M.; and Kelley, C.~T. 2000.
\newblock A locally-biased form of the DIRECT algorithm.
\newblock Technical report, North Carolina State University. Center for Research in Scientific Computation.

\bibitem[{Gao and Han(2012)}]{neldermead}
Gao, F.; and Han, L. 2012.
\newblock Implementing the Nelder-Mead Simplex Algorithm with Adaptive Parameters.
\newblock \emph{Comput. Optim. Appl.}, 51(1): 259–277.

\bibitem[{Gurobi~Optimization(2023)}]{gurobi_manual}
Gurobi~Optimization, L. 2023.
\newblock \emph{Gurobi Optimizer Reference Manual}.

\bibitem[{Hansen et~al.(2023)Hansen, yoshihikoueno, ARF1, Kadlecová, Nozawa, Rolshoven, Chan, Akimoto, brieglhostis, and Brockhoff}]{nikolaus_hansen_2023_7573532}
Hansen, N.; yoshihikoueno; ARF1; Kadlecová, G.; Nozawa, K.; Rolshoven, L.; Chan, M.; Akimoto, Y.; brieglhostis; and Brockhoff, D. 2023.
\newblock CMA-ES/pycma: r3.3.0.

\bibitem[{Henderson, Jacobson, and Johnson(2003)}]{simulatedannealing}
Henderson, D.; Jacobson, S.~H.; and Johnson, A.~W. 2003.
\newblock \emph{The Theory and Practice of Simulated Annealing}, 287--319.
\newblock Boston, MA: Springer US.

\bibitem[{Hickey, Ju, and Van~Emden(2001)}]{hickey2001interval}
Hickey, T.; Ju, Q.; and Van~Emden, M.~H. 2001.
\newblock Interval arithmetic: From principles to implementation.
\newblock \emph{Journal of the ACM (JACM)}, 48(5): 1038--1068.

\bibitem[{Jain, Kar et~al.(2017)}]{jain2017non}
Jain, P.; Kar, P.; et~al. 2017.
\newblock Non-convex optimization for machine learning.
\newblock \emph{Foundations and Trends{\textregistered} in Machine Learning}, 10(3-4): 142--363.

\bibitem[{Lavezzi, Guye, and Ciarcià(2022)}]{nonlinearbenchmark}
Lavezzi, G.; Guye, K.; and Ciarcià, M. 2022.
\newblock Nonlinear Programming Solvers for Unconstrained and Constrained Optimization Problems: a Benchmark Analysis.

\bibitem[{Liu and Lu(2014)}]{liu2014solving}
Liu, X.; and Lu, P. 2014.
\newblock Solving nonconvex optimal control problems by convex optimization.
\newblock \emph{Journal of Guidance, Control, and Dynamics}, 37(3): 750--765.

\bibitem[{Mistakidis and Stavroulakis(2013)}]{mistakidis2013nonconvex}
Mistakidis, E.~S.; and Stavroulakis, G.~E. 2013.
\newblock \emph{Nonconvex optimization in mechanics: algorithms, heuristics and engineering applications by the FEM}, volume~21.
\newblock Springer Science \& Business Media.

\bibitem[{Munos(2011)}]{DOO2011}
Munos, R. 2011.
\newblock Optimistic Optimization of a Deterministic Function without the Knowledge of its Smoothness.
\newblock In Shawe-Taylor, J.; Zemel, R.; Bartlett, P.; Pereira, F.; and Weinberger, K.~Q., eds., \emph{Advances in Neural Information Processing Systems}, volume~24. Curran Associates, Inc.

\bibitem[{Ninin(2016)}]{ninin2016global}
Ninin, J. 2016.
\newblock Global optimization based on contractor programming: An overview of the IBEX library.
\newblock In \emph{Mathematical Aspects of Computer and Information Sciences: 6th International Conference, MACIS 2015, Berlin, Germany, November 11-13, 2015, Revised Selected Papers 6}, 555--559. Springer.

\bibitem[{Olson et~al.(2012)Olson, Hashmi, Molloy, and Shehu}]{olson2012basin}
Olson, B.; Hashmi, I.; Molloy, K.; and Shehu, A. 2012.
\newblock Basin hopping as a general and versatile optimization framework for the characterization of biological macromolecules.
\newblock \emph{Advances in Artificial Intelligence}, 2012: 3--3.

\bibitem[{Pant et~al.(2020)Pant, Zaheer, Garcia-Hernandez, Abraham et~al.}]{pant2020differential}
Pant, M.; Zaheer, H.; Garcia-Hernandez, L.; Abraham, A.; et~al. 2020.
\newblock Differential Evolution: A review of more than two decades of research.
\newblock \emph{Engineering Applications of Artificial Intelligence}, 90: 103479.

\bibitem[{Puranik and Sahinidis(2017)}]{puranik2017bounds}
Puranik, Y.; and Sahinidis, N.~V. 2017.
\newblock Bounds tightening based on optimality conditions for nonconvex box-constrained optimization.
\newblock \emph{Journal of Global Optimization}, 67: 59--77.

\bibitem[{Sahinidis(2023)}]{baron_manual}
Sahinidis, N. 2023.
\newblock \emph{BARON: Global Optimization of Mixed-Integer Nonlinear Programs, {\em User's Manual}}.
\newblock The Optimization Firm, LLC, niksah@minlp.com.

\bibitem[{Storn and Price(1997)}]{storn1997differential}
Storn, R.; and Price, K. 1997.
\newblock Differential evolution-a simple and efficient heuristic for global optimization over continuous spaces.
\newblock \emph{Journal of global optimization}, 11(4): 341.

\bibitem[{Tawarmalani and Sahinidis(2005)}]{ts:05}
Tawarmalani, M.; and Sahinidis, N.~V. 2005.
\newblock {A polyhedral branch-and-cut approach to global optimization}.
\newblock \emph{Mathematical Programming}, 103: 225--249.

\bibitem[{{The Optimization Firm}(2023)}]{baron_dev}
{The Optimization Firm}. 2023.
\newblock NLP and MINLP Test Problems.
\newblock \url{https://minlp.com/nlp-and-minlp-test-problems}.
\newblock Accessed: Aug.10, 2023.

\bibitem[{Wang, Fonseca, and Tian(2020{\natexlab{a}})}]{lamcts2020}
Wang, L.; Fonseca, R.; and Tian, Y. 2020{\natexlab{a}}.
\newblock Learning Search Space Partition for Black-box Optimization using Monte Carlo Tree Search.
\newblock \emph{arXiv preprint arXiv:2007.00708}.

\bibitem[{Wang, Fonseca, and Tian(2020{\natexlab{b}})}]{wang2020learning}
Wang, L.; Fonseca, R.; and Tian, Y. 2020{\natexlab{b}}.
\newblock Learning search space partition for black-box optimization using monte carlo tree search.
\newblock \emph{Advances in Neural Information Processing Systems}, 33: 19511--19522.

\bibitem[{Xiang et~al.(2013)Xiang, Gubian, Suomela, and Hoeng}]{xiang2013generalized}
Xiang, Y.; Gubian, S.; Suomela, B.; and Hoeng, J. 2013.
\newblock Generalized simulated annealing for global optimization: the GenSA package.
\newblock \emph{R J.}, 5(1): 13.

\bibitem[{Yang(2019)}]{yang2019advancing}
Yang, T. 2019.
\newblock Advancing non-convex and constrained learning: Challenges and opportunities.
\newblock \emph{AI Matters}, 5(3): 29--39.

\bibitem[{Yanover et~al.(2006)Yanover, Meltzer, Weiss, Bennett, and Parrado-Hern{\'a}ndez}]{yanover2006linear}
Yanover, C.; Meltzer, T.; Weiss, Y.; Bennett, K.~P.; and Parrado-Hern{\'a}ndez, E. 2006.
\newblock Linear Programming Relaxations and Belief Propagation--An Empirical Study.
\newblock \emph{Journal of Machine Learning Research}, 7(9).

\bibitem[{Zhai and Gao(2022)}]{zhaimonte}
Zhai, Y.; and Gao, S. 2022.
\newblock Monte Carlo Tree Descent for Black-Box Optimization.
\newblock In \emph{Advances in Neural Information Processing Systems}.

\bibitem[{Zhu et~al.(1997)Zhu, Byrd, Lu, and Nocedal}]{zhu1997algorithm}
Zhu, C.; Byrd, R.~H.; Lu, P.; and Nocedal, J. 1997.
\newblock Algorithm 778: L-BFGS-B: Fortran subroutines for large-scale bound-constrained optimization.
\newblock \emph{ACM Transactions on mathematical software (TOMS)}, 23(4): 550--560.

\end{thebibliography}

\section{Appendix}

\subsection{Nomenclature}
Following is the modified UCT criterion coupled with the function approximation used in MCIR, as described in Eq.~\ref{eq:uct}.
\begin{equation*}
\label{eq:uct_appendix}
u(n_{ci}) = -y^*_{ci} 
       - C_{lb} \cdot lb_{ci} 
       - C_{v} \cdot V_{ci}
       + C_{x}  \cdot \sqrt{\frac{\log{N_{p}}}{N_{ci}}}
\end{equation*}
In this section we summarize the symbols and terms used in MCIR (Alg.\ref{alg:MCIR}).

\begin{itemize}
    \item Node $n$: we use character $n$ to represent a node; subscripts are used to distinguish different node with its id: e.g., $n_p$ is a parent node, and $\{n_{ci}\}$ are a set of child nodes, for $i=1, ...$

    \item Best found sample $(x, y^*)$: a node $n$ is represented by its best-found sample $(x, y^*)$, which is the sample with lowest objective function value found on it and its subtree. On node $n_p$, we label the best found sample input vector as $x_p$, and corresponding function $y^*_p$. Similarly, for nodes $\{n_{ci}\}$, $i=1,...$, their best found sample input vectors and function values are $\{x_{ci}\}$ and $\{y^*_{ci}\}$, respectively.
    \item Box $B$: for a node $n$, we assign a box $B \subseteq \Omega $ as the bounds in the input domain. We use $B_p$ to indicate the box assigned to the node $n_p$.
    \item Function value lower bound $lb$: The function value interval on an input box $B$ is computed as $f(B) = [lb(f(B)), ub(f(B))]$. We use $lb_{p}$ to indicate $lb(f(B_{p}))$.
    \item $V_p$ is used to denote the volume of the box $B_p$.
    \item $N$ is the number of visits to the node
    \item $C_{lb}$ is the weight factor that controls the importance of the function's lower bound in Eq.~\ref{eq:uct}
    \item $C_{v}$ is the weight factor associated with the volume of the box where the lower bound is identified. 
    \item $C_{x}$ is the hyper-parameter for the extent of visitation-based exploration. 
\end{itemize}

\subsection{Experiment Details}
\subsubsection{Test Sets}
The test sets comprise functions from three different categorises:
synthetic functions designed for nonlinear optimization, 
bounded-constrained non-convex global optimization problems derived from real-world scenarios, 
and neural networks fitting for single valued functions. 

\begin{figure*}[t!]
\centering    
\subfloat[Ackley-2d]{\label{fig:ls_ackley}\includegraphics[width=0.32\textwidth]{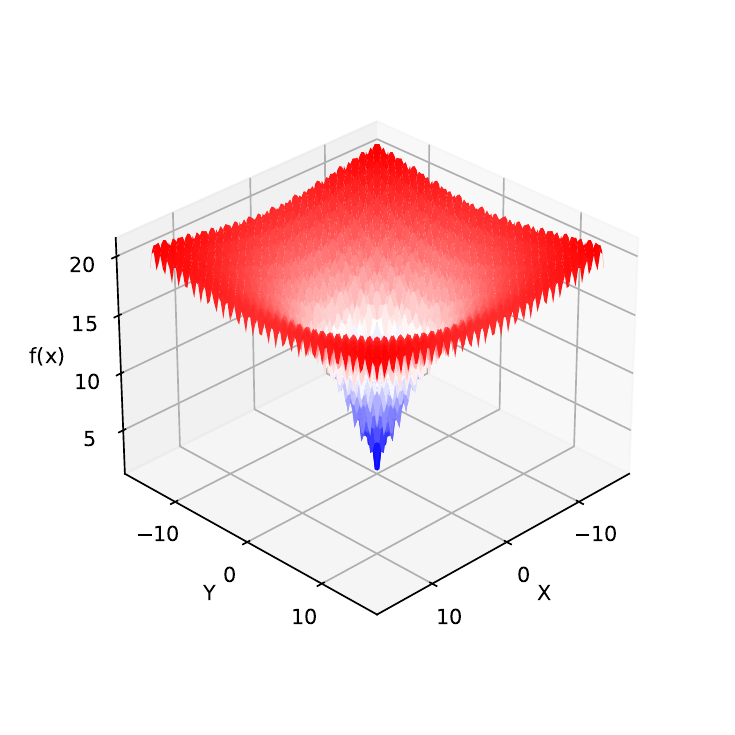}}
\subfloat[Levy-2d]{\label{fig:ls_levy}\includegraphics[width=0.32\textwidth]{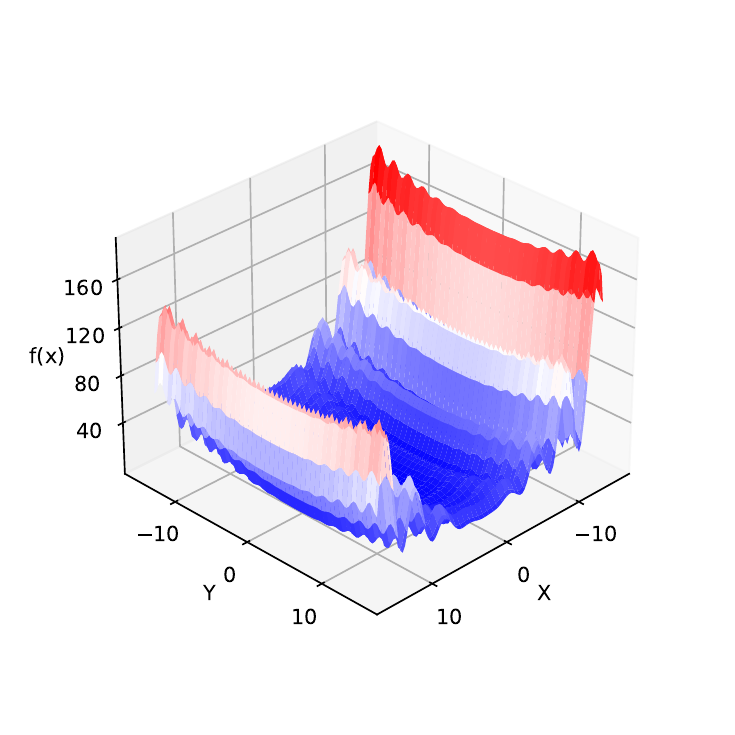}}
\subfloat[Michalewicz-2d]{\label{fig:ls_mich}\includegraphics[width=0.32\textwidth]{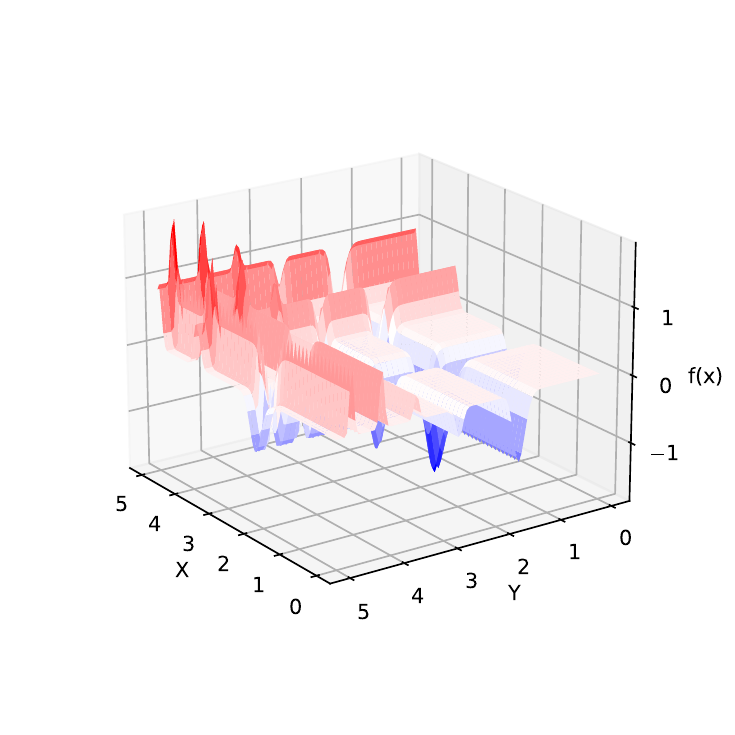}}
\caption{Landscape of test functions: Ackley, Levy, and Michalewicz.}
\label{fig:landscape}
\end{figure*}

Synthetic functions \cite{nonlinearbenchmark} are widely-used in nonlinear optimization benchmarks .
We choose three functions: \textbf{Levy}, \textbf{Ackley}, and \textbf{Michalewicz}, and examine our algorithms performances on the functions in 50d, 100d, and 200d. 
The Levy function (Fig.~\ref{fig:ls_levy}) is known for its complex, multi-modal landscape, featuring numerous local optima separated by vast valleys. 
The Ackley function (Fig.~\ref{fig:ls_ackley}) presents a rugged landscape with a prominent, flat region surrounding the global minimum. 
The Michalewicz function (Fig.~\ref{fig:ls_mich}) is characterized by its highly oscillatory and irregular landscape, characterized by numerous peaks and valleys. 

We use three test functions from bounded-constrained non-convex problem sets for global optimization \cite{puranik2017bounds}: \textbf{Biggsbi1} (1000d), \textbf{Harkerp} (100d), and \textbf{Watson} (31d).
Biggsbil1 is a function modeling the kinetics of a biochemical reaction. It has a global minimum that's surrounded by several local minima making it challenging for optimization algorithms.
Harkerp is a mathematical problem in the field of economics, specifically in the area of profit maximization. 
This non-convex problem simulates a firm seeking to maximize profit through optimal pricing and advertising decisions, subject to market demand and cost constraints. 
Watson is a smooth, non-convex function known for its narrow, curved valley that leads to the global minimum.

Finally, we ventured into the realm of neural network optimization. 
Our methodology enabled us to transform a network with ReLU activations function into an analytic expression. 
This, however, introduces challenges due to the nonlinearity arising from ReLU activations and the partitioning of the input space. 
The global minimum of neural networks becomes elusive, especially due to the unpredictable error. 
Moreover, the node learning step in our algorithm can no longer use Hessian information because the Hessian of ReLU-activated neural network is zero.
To balance computational cost and evaluation complexity, we trained the neural network to fit the Ackley and Michalewicz functions in 50d and 100d spaces with 16 hidden neurons in a single layer.

\subsubsection{Hyperparameters}

Tab.~\ref{tab:hyper-parameter} presents the hyperparameters utilized for benchmarking MCIR across various functions. 
In this context, the notation \textbf{\# CN} corresponds to the quantity of appended child nodes during expansion. Notably, a higher count of child nodes contributes to an increased number of samples dedicated to exploiting the selected leaf node. 
Meanwhile, \textbf{\# LO} denotes the number of samples permissible for the local optimizer to undertake in enhancing the sample quality on the node. 
We intentionally constrain this number to a low value, thereby preventing an excessive reliance on the local optimizer.

\begin{table}[]
    \centering
    \begin{tabular}{|l|c|c|c|c|c|}
    \hline
    \textbf{Functions} & $\mathbf{C}_{lb}$ 
    & $\mathbf{C}_{v}$ & $\mathbf{C}_{x}$ 
    & \# CN & \# LO \\
    \hline
    \textbf{Ackley-50d} & 50.0 & 0.5 & 0.5 & 10 & 50 \\ 
    \textbf{Ackley-100d} & 50.0 & 0.5 & 0.5 & 20 & 50 \\ 
    \textbf{Ackley-200d} & 50.0 & 0.5 & 0.5 & 30 & 50 \\ 
    \hline 
    \textbf{Levy-50d} & 50.0 & 0.5 & 0.5 & 10 & 50 \\ 
    \textbf{Levy-100d} & 50.0 & 0.5 & 0.5 & 20 & 50 \\ 
    \textbf{Levy-200d} & 50.0 & 0.5 & 0.5 & 30 & 50 \\ 
    \hline
    \textbf{Micha.-50d} & 50.0 & 0.5 & 0.5 & 10 & 20 \\ 
    \textbf{Micha.-100d} & 50.0 & 0.5 & 0.5 & 20 & 50 \\ 
    \textbf{Micha.-100d} & 50.0 & 0.5 & 0.5 & 30 & 50 \\     
    \hline
    \textbf{Biggsbi1-1000d} & 50.0 & 0.5 & 0.5 & 10 & 10 \\
    \textbf{Harkerp-100d} & 50.0 & 0.5 & 0.5 & 10 & 10 \\
    \textbf{Harkerp-31d} & 50.0 & 0.5 & 0.5 & 10 & 10 \\
    \hline
    \textbf{NN-Ackley-50d} & 50.0 & 0.5 & 0.5 & 10 & 20 \\ 
    \textbf{NN-Ackley-100d} & 50.0 & 0.5 & 0.5 & 20 & 50 \\ 
    \textbf{NN-Micha.-50d} & 50.0 & 0.5 & 0.5 & 10 & 20 \\ 
    \textbf{NN-Micha.-100d} & 50.0 & 0.5 & 0.5 & 20 & 50 \\ 
    \hline
    \end{tabular}
    \caption{Hyperparameters used for MCIR}
    \label{tab:hyper-parameter}
\end{table}

Hyperparameters for other baseline:
\begin{itemize}
    \item \textbf{Basinhopping}: $niter=1000$, $niter\_success=1000$ 
    \item \textbf{Differential Evolution}: $maxiter=1000000$
    \item \textbf{Dual Annealing}: \textit{no local search=True}
    \item \textbf{Direct}: $maxiter=100000$, $locally\_biased=False$
    \item \textbf{CMA}: use $fmin2$, with $\sigma = 1$ 
    \item \textbf{TuRBO}: $n\_init=20$, $batch\_size=40$, $use\_ard=True$, $n\_training\_steps=50$
    \item \textbf{LaMCTS}: $n\_init=50$, $C_p = 2.0$, $leaf\_size=50$
\end{itemize}

Hyperparameters not specifically addressed above were subjected with their default values within the packages.

\subsubsection{Benchmark Results}
In this section, we use the abbreviations as follows to denote different optimization algorithms:

\begin{itemize}
\item \textbf{BH}: Basinhopping
\item \textbf{DE}: Differential Evolution
\item \textbf{DIRECT}: DIRECT
\item \textbf{DA}: Dual Annealing
\item \textbf{CMA}: CMA
\item \textbf{TuRBO}: TuRBO
\item \textbf{LaMCTS}: LaMCTS
\item \textbf{Gurobi}: the Gurobi solver
\item \textbf{MCIR}: our algorithm, MCIR
\end{itemize}

Tab.~\ref{tab:result_sheet_ackley_levy}, Tab.~\ref{tab:result_sheet_Mich}, and Tab.~\ref{tab:result_sheet_bcp_nn} provide detailed benchmark results.
The column labeled \textbf{$y^* \pm std$} represents the best-found objective function value. 
The column labeled \textbf{Time}$^*$ denotes the clock time, in seconds, at which the algorithm found the best point for the first time. 
The column labeled \textbf{\# of Samples$^*$} indicates the number of samples which have been evaluated when the algorithm reached the best-found point for the first time.

In the case of \textbf{Gurobi}, our testing was exclusively conducted on functions that could be precisely and accurately formulated. 
Consequently, its presence might be limited.
Furthermore, in the cases where we carefully formulated the problems, Gurobi can solve the problem at pre-processing stage resulting in a zero seconds computational time and zero times of function evaluation calls.

For \textbf{TuRBO} and \textbf{LaMCTS}, the number of function evaluations are still quite limited, even after we have increased the timeout as five times longer than other baselines.
Particularly for \textbf{LaMCTS}, there are only hundreds of function evaluations after 5 hours of computational time.
Therefore, it is hard to determine when they first encountered their best-found sample.

\begin{table*}[]
    \centering
    \begin{tabular}{|c|l|c|c|c|}
    \hline
    \textbf{Function} & \textbf{Alg.} &
    \textbf{$y^* \pm std $} & \textbf{Time$^*$} &
    \textbf{\# Samples $^*$} \\
    \hline
\textbf{Ackley-50d}
& BH & $20.8358 \pm 0.1967 $ & $1.47 \pm 0.62 $ & $12709 \pm 5240 $  \\ 
& DE & $20.1856 \pm 0.2135 $ & $0.45 \pm 0.16 $ & $1171 \pm 170 $  \\ 
& DIRECT & $4.5525  $ & $0.00  $ & $0  $  \\ 
& DA & $0.0011 \pm 0.0001 $ & $110.75 \pm 15.27 $ & $123591 \pm 1364 $  \\ 
 & CMA &$0.0000 \pm 0.0000$ & $5.68 \pm 0.29$ & $6792 \pm 375$ \\
& TuRBO &$2.0531 \pm 0.0956$ &  & $8022 \pm 3248$ \\
& LaMCTS &$20.5670 \pm 0.3844$ & & $170 \pm 220$ \\
& Gurobi & $2.0217 $ & $0.00 $ & $0  $  \\ 
& MCIR & $0.6245 \pm 0.7649 $ & $118.12 \pm 15.51 $ & $23529 \pm 3389 $  \\ 
\hline
\textbf{Ackley-100d}
& BH & $20.9803 \pm 0.1058 $ & $1.61 \pm 1.04 $ & $14342 \pm 9265 $  \\ 
& DE & $20.6655 \pm 0.0568 $ & $0.82 \pm 0.19 $ & $2328 \pm 689 $  \\ 
& DIRECT & $4.5525 \pm 0.0000 $ & $0.00 \pm 0.00 $ & $0 $  \\ 
& DA & $0.0021 \pm 0.0002 $ & $287.41 \pm 19.83 $ & $265813 \pm 4155 $  \\ 
& CMA &$0.0000 \pm 0.0000$ & $10.09 \pm 0.20$ & $12226 \pm 244$ \\
& TuRBO &$5.1832 \pm 0.5763$ &  & $6344 \pm 105$ \\
& LaMCTS &$20.8962 \pm 0.0350$ &  & $249 \pm 136$ \\
& Gurobi & $2.0217 \pm 0.0000 $ & $0.00  $ & $0  $  \\ 
& MCIR & $0.0000 \pm 0.0000 $ & $1033.78 \pm 55.01 $ & $61351 \pm 1469 $  \\ 
\hline
\textbf{Ackley-200d}
& BH & $21.1346 \pm 0.0700 $ & $2.29 \pm 2.24 $ & $20183 \pm 19168 $  \\ 
& DE & $20.6334 \pm 0.1212 $ & $1.76 \pm 0.25 $ & $4790 \pm 976 $  \\ 
& DIRECT & $4.5525 \pm 0.0000 $ & $0.00 $ & $0 $  \\ 
& DA & $0.0041 \pm 0.0001 $ & $766.03 \pm 51.73 $ & $546779 \pm 7893 $  \\
& CMA &$0.0000 \pm 0.0000$ & $17.91 \pm 1.08$ & $23012 \pm 1678$ \\
& TuRBO &$14.2829 \pm 0.4982$ &  & $5695 \pm 131$ \\
& LaMCTS &$21.0537 \pm 0.0283$ &  & $223 \pm 106$ \\
& Gurobi & $2.0217 \pm 0.0000 $ & $0.00 $ & $0 $  \\ 
& MCIR & $0.5135 \pm 1.0270 $ & $6544.56 \pm 1661.44 $ & $114651 \pm 23828 $  \\ 
\hline
\textbf{Levy-50d}
& BH & $114.9944 \pm 18.8022 $ & $44.35 \pm 4.47 $ & $98070 \pm 12008 $  \\ 
& DE & $0.1373 \pm 0.1733 $ & $288.61 \pm 155.92 $ & $450636 \pm 232094 $  \\ 
& DIRECT & $13.7759 \pm 0.0000 $ & $0.20 \pm 0.05 $ & $199 $  \\ 
& DA & $0.0000 \pm 0.0000 $ & $86.62 \pm 30.43 $ & $59073 \pm 24203 $  \\ 
& CMA &$0.2149 \pm 0.1561$ & $10.94 \pm 3.96$ & $16597 \pm 6516$ \\
& TuRBO &$5.4087 \pm 3.0669$ &  & $8309 \pm 3148$ \\
& LaMCTS &$84.7584 \pm 7.0483$ & & $648 \pm 87$ \\
& MCIR & $5.5889 \pm 1.9306 $ & $168.77 \pm 101.57 $ & $13469 \pm 8078 $  \\ 
\hline
\textbf{Levy-100d}
& BH & $248.9395 \pm 32.5442 $ & $101.83 \pm 4.58 $ & $206809 \pm 16550 $  \\ 
& DE & $3.1676 \pm 0.6679 $ & $791.58 \pm 372.37 $ & $778528 \pm 59597 $  \\ 
& DIRECT & $27.4119 \pm 0.0000 $ & $0.28 \pm 0.06 $ & $399 $  \\ 
& DA & $0.0001 \pm 0.0002 $ & $211.72 \pm 27.53 $ & $138905 \pm 10619 $  \\ 
& CMA &$2.0102 \pm 0.8746$ & $18.74 \pm 13.50$ & $31149 \pm 24441$ \\
& TuRBO &$15.7115 \pm 6.1465$ &  & $5501 \pm 177$ \\
& LaMCTS &$326.9397 \pm 38.7379$ & & $667 \pm 43$ \\
& Gurobi & $-0.0000 \pm 0.0000 $ & $0.00  $ & $0 $  \\ 
& MCIR & $9.1970 \pm 1.4018 $ & $1810.49 \pm 1180.24 $ & $35611 \pm 23348 $  \\ 
\hline
\textbf{Levy-200d}
& BH & $660.9173 \pm 51.2398 $ & $187.51 \pm 19.31 $ & $377479 \pm 35511 $  \\ 
& DE & $23.5931 \pm 4.9629 $ & $1714.29 \pm 598.96 $ & $1909534 \pm 265067 $  \\ 
& DIRECT & $54.6839 \pm 0.0000 $ & $0.52 \pm 0.09 $ & $799 $  \\ 
& DA & $0.0005 \pm 0.0005 $ & $639.01 \pm 89.80 $ & $409497 \pm 58671 $  \\ 
& CMA &$7.5793 \pm 1.5815$ & $28.14 \pm 16.21$ & $47511 \pm 37708$ \\
& TuRBO &$169.9204 \pm 13.7658$ &  & $5591 \pm 108$ \\
& LaMCTS &$1533.9164 \pm 111.0211$ & & $504 \pm 93$ \\
& Gurobi & $-0.0002 \pm 0.0000 $ & $0.00 $ & $0 $  \\ 
& MCIR & $20.5379 \pm 4.1395 $ & $5388.18 \pm 348.55 $ & $45079 \pm 1564 $  \\ 
\hline

    \end{tabular}
    \caption{Benchmark results for Ackley and Levy}
    \label{tab:result_sheet_ackley_levy}
\end{table*}

\begin{table*}[]
    \centering
    \begin{tabular}{|c|l|c|c|c|}
    \hline
    \textbf{Function} & \textbf{Alg.} &
    \textbf{$y^* \pm std $} & \textbf{Time$^*$} &
    \textbf{\# Samples $^*$} \\
    \hline
\textbf{Michalewicz-50d}
& BH & $-8.8432 \pm 0.8271 $ & $55.45 \pm 76.27 $ & $209182 \pm 185755 $  \\ 
& DE & $-20.1906 \pm 1.0243 $ & $212.15 \pm 93.79 $ & $751515 \pm 644 $  \\ 
& DIRECT & $-13.0827 \pm 0.0000 $ & $1.98 \pm 0.07 $ & $25609 $  \\ 
& DA & $-49.5421 \pm 0.0134 $ & $88.54 \pm 7.28 $ & $127056 \pm 2375 $  \\
& CMA &$-22.9280 \pm 3.5200$ & $26.95 \pm 4.91$ & $42784 \pm 9507$ \\
& TuRBO &$-35.3835 \pm 1.1419$ &  & $15915 \pm 11657$ \\
& LaMCTS &$-13.9431 \pm 0.6567$ & & $463 \pm 291$ \\
& Gurobi & $-49.8521 \pm 0.0000 $ & $0.00 $ & $0 $  \\ 
& MCIR & $-48.0423 \pm 0.7511 $ & $177.66 \pm 1.97 $ & $25418 \pm 283 $  \\ 
\hline
\textbf{Michalewicz-100d}
& BH & $-14.6945 \pm 0.5995 $ & $15.15 \pm 12.32 $ & $89181 \pm 75509 $  \\ 
& DE & $-30.6034 \pm 0.7476 $ & $702.97 \pm 338.89 $ & $1503114 \pm 1322 $  \\ 
& DIRECT & $-26.1498 \pm 0.0000 $ & $35.85 \pm 3.22 $ & $103813 $  \\ 
& DA & $-99.4230 \pm 0.0207 $ & $228.55 \pm 28.25 $ & $297807 \pm 5504 $  \\ 
& CMA &$-34.9603 \pm 3.7703$ & $64.33 \pm 10.81$ & $111835 \pm 21423$ \\
& TuRBO &$-51.6695 \pm 1.9650$ &  & $27427 \pm 9051$ \\
& LaMCTS &$-23.8377 \pm 1.1338$ & & $522 \pm 197$ \\
& Gurobi & $-100.5415 \pm 0.0000 $ & 0.0  & 0  \\ 
& MCIR & $-91.9268 \pm 1.0598 $ & $3518.56 \pm 69.82 $ & $96317 \pm 2286 $  \\ 
\hline
\textbf{Michalewicz-200d}
& BH & $-27.3023 \pm 1.6926 $ & $3394.23 \pm 1876.30 $ & $1649919 \pm 870987 $  \\ 
& DE & $-48.4502 \pm 2.0750 $ & $1807.93 \pm 365.10 $ & $3008602 \pm 4315 $  \\ 
& DIRECT & $-51.1477 \pm 0.0000 $ & $29.64 \pm 6.27 $ & $205917 $  \\ 
& DA & $-199.1417 \pm 0.0593 $ & $503.03 \pm 60.18 $ & $766905 \pm 11713 $  \\ 
& CMA &$-43.1354 \pm 10.5304$ & $155.91 \pm 78.58$ & $310636 \pm 156979$ \\
& TuRBO &$-96.4577 \pm 2.6720$ &  & $8502 \pm 3394$ \\
& LaMCTS &$-39.4668 \pm 1.3005$ & & $355 \pm 171$ \\
& Gurobi & $-202.4635 \pm 0.0000 $ & $0.00  $ & $0 $  \\ 
& MCIR & $-172.5783 \pm 2.8443 $ & $7146.98 \pm 167.62 $ & $51914 \pm 856 $  \\ 
    \hline
    \end{tabular}
    \caption{Benchmark results for Michalewicz}
    \label{tab:result_sheet_Mich}
\end{table*}

\begin{table*}[]
    \centering
    \begin{tabular}{|c|l|c|c|c|}
    \hline
    \textbf{Function} & \textbf{Alg.} &
    \textbf{$y^* \pm std $} & \textbf{Time$^*$} &
    \textbf{\# Samples $^*$} \\
    \hline
\textbf{Biggsbi1-1000d}
& DE & $56.1697 \pm 3.8586 $ & $3555.39 \pm 466.69 $ & $232118 \pm 28533 $  \\ 
& DIRECT & $1.6592 \pm 0.0000 $ & $4.60 \pm 0.65 $ & $4001 $  \\ 
& DA & $7.5562 \pm 2.9694 $ & $3615.67 \pm 14.28 $ & $362556 \pm 34168 $  \\ 
& CMA &$72.3690 \pm 3.7933$ & $166.55 \pm 5.90$ & $448919 \pm 10571$ \\
& TuRBO &$32.4614 \pm 1.5262$ &  & $6649 \pm 22$ \\
& LaMCTS &$110.3565 \pm 6.6033$ & & $346 \pm 129$ \\
& MCIR & $1.0101 \pm 0.0001 $ & $1986.40 \pm 277.34 $ & $17584 \pm 2360 $  \\ 
\hline
\textbf{Harkerp-100d}
& DE & $-0.7887 \pm 0.0738 $ & $2214.26 \pm 1337.87 $ & $1112523 \pm 514176 $  \\ 
& DIRECT & $5266895.7760 \pm 0.0000 $ & $525.71 \pm 955.15 $ & $86592 $  \\ 
& DA & $-0.6831 \pm 0.1248 $ & $23.94 \pm 30.71 $ & $15265 \pm 60 $  \\ 
& CMA &$-0.3220 \pm 0.2179$ & $34.24 \pm 27.89$ & $61240 \pm 50743$ \\
& TuRBO &$91484.0598 \pm 18502.6283$ &  & $6251 \pm 39$ \\
& LaMCTS &$980949.0111 \pm 122103.2066$ & & $520 \pm 24$ \\
& MCIR & $-0.9256 \pm 0.0017 $ & $236.05 \pm 60.83 $ & $13106 \pm 3312 $  \\ 
\hline
\textbf{Watson-31d}
& DE & $0.0010 \pm 0.0006 $ & $228.30 \pm 39.65 $ & $492335 \pm 2941 $  \\ 
& DIRECT & $264.6848 \pm 0.0000 $ & $0.90 \pm 0.04 $ & $1849 $  \\ 
& DA & $0.1415 \pm 0.1318 $ & $3.58 \pm 0.54 $ & $7665 \pm 102 $  \\ 
& CMA &$0.0000 \pm 0.0000$ & $26.68 \pm 3.03$ & $42245 \pm 6335$ \\
& TuRBO &$4.6289 \pm 1.8125$ &  & $16500 \pm 14087$ \\
& LaMCTS &$4109.8301 \pm 1959.8004$ & & $97 \pm 36$ \\
& MCIR & $0.0023 \pm 0.0018 $ & $487.88 \pm 288.47 $ & $43682 \pm 25967 $  \\ 
\hline

\textbf{NN-Ackley-50d}
& BH & $21.1501 \pm 0.1014 $ & $29.95 \pm 22.82 $ & $17839 \pm 14101 $  \\ 
& DE & $21.0200 \pm 0.0181 $ & $2.09 \pm 0.94 $ & $1056 \pm 466 $  \\ 
& DIRECT & $20.8103 \pm 0.0000 $ & $56.48 \pm 1.82 $ & $36401 $  \\ 
& DA & $20.7037 \pm 0.0009 $ & $195.30 \pm 2.30 $ & $108967 \pm 1559 $  \\
& CMA &$20.7056 \pm 0.0030$ & $17.88 \pm 5.20$ & $26648 \pm 9448$ \\
& TuRBO &$20.7524 \pm 0.0023$ &  & $27706 \pm 18354$ \\
& LaMCTS &$21.1075 \pm 0.0250$ & & $150 \pm 156$ \\
& MCIR & $20.7481 \pm 0.0152 $ & $598.52 \pm 391.36 $ & $19031 \pm 14205 $  \\ 
\hline
\textbf{NN-Ackley-100d}
& BH & $21.2368 \pm 0.0307 $ & $51.32 \pm 26.29 $ & $30847 \pm 15151 $  \\ 
& DE & $21.1453 \pm 0.0080 $ & $3.49 \pm 1.30 $ & $1723 \pm 645 $  \\ 
& DIRECT & $21.0549 \pm 0.0000 $ & $116.69 \pm 6.78 $ & $72091 $  \\ 
& DA & $20.9845 \pm 0.0010 $ & $389.19 \pm 18.13 $ & $218588 \pm 3637 $  \\ 
& CMA &$20.9828 \pm 0.0020$ & $53.63 \pm 15.49$ & $95970 \pm 33576$ \\
& TuRBO &$21.0412 \pm 0.0030$ &  & $21576 \pm 13880$ \\
& LaMCTS &$21.1735 \pm 0.0111$ & & $122 \pm 128$ \\
& MCIR & $21.0162 \pm 0.0139 $ & $2460.72 \pm 426.62 $ & $39661 \pm 2727 $  \\ 
\hline
\textbf{NN-Michalewicz-50d}
& DE & $-7.8478 \pm 0.6904 $ & $151.45 \pm 108.62 $ & $79255 \pm 57812 $  \\ 
& DIRECT & $-6.6457 \pm 0.0000 $ & $1.48 \pm 0.20 $ & $771 $  \\ 
& DA & $-7.6015 \pm 0.6065 $ & $224.88 \pm 19.00 $ & $129478 \pm 12950 $  \\ 
& CMA &$-7.2868 \pm 0.3838$ & $25.40 \pm 8.30$ & $42098 \pm 16777$ \\
& TuRBO &$-8.2827 \pm 0.0133$ &  & $22526 \pm 10821$ \\
& LaMCTS &$-6.8803 \pm 0.0271$ & & $755 \pm 120$ \\
& MCIR & $-9.6448 \pm 0.0384 $ & $711.72 \pm 463.97 $ & $25225 \pm 16613 $  \\ 
\hline
\textbf{NN-Michalewicz-100d}
& DE & $-12.8481 \pm 1.1304 $ & $115.02 \pm 171.16 $ & $57962 \pm 86966 $  \\ 
& DIRECT & $-12.4153 \pm 0.0000 $ & $3.93 \pm 0.36 $ & $2391 $  \\ 
& DA & $-14.6488 \pm 2.4041 $ & $430.69 \pm 29.31 $ & $261728 \pm 27258 $  \\ 
& CMA &$-12.7314 \pm 0.0352$ & $63.74 \pm 24.53$ & $120090 \pm 55975$ \\
& TuRBO &$-15.1034 \pm 0.2637$ &  & $26886 \pm 18835$ \\
& LaMCTS &$-12.2140 \pm 0.1088$ & & $573 \pm 135$ \\
& MCIR & $-17.5062 \pm 0.0068 $ & $6060.88 \pm 4328.83 $ & $101634 \pm 72926 $  \\ 
\hline
    \end{tabular}
    \caption{Benchmark results for bounded-constrained functions and NN functions}
    \label{tab:result_sheet_bcp_nn}
\end{table*}

\end{document}